\documentclass[letterpaper]{article} % DO NOT CHANGE THIS
\usepackage{aaai2026}  % DO NOT CHANGE THIS
\usepackage{times}  % DO NOT CHANGE THIS
\usepackage{helvet}  % DO NOT CHANGE THIS
\usepackage{courier}  % DO NOT CHANGE THIS
\usepackage[hyphens]{url}  % DO NOT CHANGE THIS
\usepackage{graphicx} % DO NOT CHANGE THIS
\usepackage{natbib}  % DO NOT CHANGE THIS AND DO NOT ADD ANY OPTIONS TO IT
\usepackage{caption} % DO NOT CHANGE THIS AND DO NOT ADD ANY OPTIONS TO IT
\usepackage{algorithm}
\usepackage{algorithmic}
\usepackage{amsmath} % for advanced math formatting
\usepackage{amssymb} % for symbols like \mathbb{R}
\usepackage{bbding}
\usepackage{xcolor}
\usepackage{multirow}
\usepackage{graphicx} % For resizebox
\usepackage{amsmath} % For math symbols like \uparrow
\usepackage{xcolor} % For colors if needed
\usepackage{tabularray} % The main table package
\usepackage{booktabs} % For professional-looking horizontal rules

\usepackage{newfloat}
\usepackage{listings}
\DeclareCaptionStyle{ruled}{labelfont=normalfont,labelsep=colon,strut=off} % DO NOT CHANGE THIS
\floatstyle{ruled}
\newfloat{listing}{tb}{lst}{}
\floatname{listing}{Listing}
\usepackage{xspace}

\usepackage{amsmath}

\usepackage{subfigure}
\usepackage{subfig}
\usepackage{multirow}
\usepackage[table,xcdraw]{xcolor} 
\usepackage{booktabs}
\usepackage{adjustbox}
\usepackage{pifont}
\usepackage{graphicx}
\usepackage{subcaption}
\usepackage{amssymb}
\usepackage[hyphens]{url}  % DO NOT CHANGE THIS
\title{MMMamba: A Versatile Cross-Modal In Context Fusion Framework for Pan-Sharpening and Zero-Shot Image Enhancement}
\author{
    Yingying Wang\textsuperscript{\rm 1}\equalcontrib,
    Xuanhua He\textsuperscript{\rm 2}\equalcontrib,
    Chen Wu\textsuperscript{\rm 3}\equalcontrib,
    Jialing Huang\textsuperscript{\rm 1},
    Suiyun Zhang\textsuperscript{\rm 4},
    Rui Liu\textsuperscript{\rm 4},
    \\ Xinghao Ding\textsuperscript{\rm 1}, 
    Haoxuan Che\textsuperscript{\rm 4}\thanks{Corresponding Author.}
}
\affiliations{
     \textsuperscript{\rm 1} Key Laboratory of Multimedia Trusted Perception and Efficient Computing,\\ Ministry of Education of China, Xiamen University, China\\
     \textsuperscript{\rm 2}The Hong Kong University of Science and Technology\\
     \textsuperscript{\rm 3}University of Science and Technology of China\\
     \textsuperscript{\rm 4}Huawei Research\\
    
    {wangyingying7@stu.xmu.edu.cn, xhecd@connect.ust.hk, wuchen5x@mail.ustc.edu.cn, che.haoxuan@huawei.com}\\
}

\usepackage{bibentry}
\begin{document}

\maketitle

\begin{abstract}
Pan-sharpening aims to generate high-resolution multispectral (HRMS) images by integrating a high-resolution panchromatic (PAN) image with its corresponding low-resolution multispectral (MS) image. To achieve effective fusion, it is crucial to fully exploit the complementary information between the two modalities. Traditional CNN-based methods typically rely on channel-wise concatenation with fixed convolutional operators, which limits their adaptability to diverse spatial and spectral variations. While cross-attention mechanisms enable global interactions, they are computationally inefficient and may dilute fine-grained correspondences, making it difficult to capture complex semantic relationships. Recent advances in the Multimodal Diffusion Transformer (MMDiT) architecture have demonstrated impressive success in image generation and editing tasks. Unlike cross-attention, MMDiT employs in-context conditioning to facilitate more direct and efficient cross-modal information exchange. In this paper, we propose MMMamba, a cross-modal in-context fusion framework for pan-sharpening, with the flexibility to support image super-resolution in a zero-shot manner. Built upon the Mamba architecture, our design ensures linear computational complexity while maintaining strong cross-modal interaction capacity. Furthermore, we introduce a novel multimodal interleaved (MI) scanning mechanism that facilitates effective information exchange between the PAN and MS modalities. Extensive experiments demonstrate the superior performance of our method compared to existing state-of-the-art (SOTA) techniques across multiple tasks and benchmarks. 
\end{abstract}
\begin{links}
\link{Code}{https://github.com/Gracewangyy/MMMamba}
\end{links}


\section{Introduction}
With the growing demand for high-quality satellite imagery in areas such as agriculture \cite{jenerowicz2016pan}, urban planning \cite{aiazzi2003mtf}, and environmental monitoring \cite{sunuprapto2016evaluation}, obtaining high-resolution multi-spectral (HRMS) data has become more critical than ever. However, the physical limitations of satellite sensors impede the direct acquisition of multi-spectral images that offer both fine spatial detail and rich spectral information. To address this issue, most satellites are equipped with two separate sensors: panchromatic (PAN) and multi-spectral (MS), each designed to capture complementary aspects. PAN images provide high spatial resolution but limited spectral coverage, while MS images offer rich spectral information at lower spatial resolutions. Pan-sharpening has therefore emerged as a practical and essential technique, aiming to fuse these two data sources into a single image that combines the spatial sharpness of PAN with the spectral fidelity of MS.

Early efforts in pan-sharpening were predominantly based on classical paradigms such as component substitution (CS) \cite{kwarteng1989extracting}, multi-resolution analysis (MRA) \cite{mallat2002theory}, and variational optimization (VO) \cite{ballester2006variational}. These hand-crafted techniques relied on physical modeling and prior domain knowledge, limiting their ability to capture complex cross-modal relationships and yielding suboptimal results. The introduction of deep learning into the pan-sharpening field has led to significant improvements in both spatial resolution and spectral fidelity. A notable breakthrough was the pioneering PNN model \cite{masi2016pansharpening}, which demonstrated remarkable performance improvements over traditional approaches. Since then, the research community has witnessed rapid advancements with increasingly sophisticated neural network architectures \cite{wang2025learning, li2025freq}. Based on varying fusion strategies, these approaches can be broadly categorized into channel concatenation-based methods, such as DIRFL \cite{lin2023domain} and HFEAN \cite{wang2023learning}, PAN injection with multi-scale techniques like MSDDN \cite{he2023multiscale} and WaveletNet \cite{WaveletNet}, cross-attention methods exemplified by Panformer \cite{zhou2022panformer} and CMINet \cite{wang2024cross}, and gating-based approaches, including FAME \cite{he2024frequency} and Pan-Mamba \cite{he2025pan}.

Despite their progress, existing methods still exhibit certain limitations that impede further performance improvements. CNN-based approaches typically rely on channel-wise concatenation, a static mechanism that lacks the adaptive flexibility to model the complex relationships between modalities. Transformer-based methods, while employing cross-attention and offering more dynamism, still have their drawbacks. First, they aggregate features through weighted averaging, which tends to smooth out the high-frequency spatial details crucial for preserving the integrity of the PAN image. Second, the information flows in only one direction, restricting the depth and richness of the interaction between modalities. Recent architectures, such as the Multimodal Diffusion Transformer (MMDiT) \cite{esser2024scaling}, have demonstrated significant success in multimodal interaction by adopting an in-context conditioning strategy \cite{tan2024ominicontrol, labs2025flux}. This approach discards traditional fusion modules like channel concatenation and cross-attention, instead concatenating tokens from all modalities into a single unified sequence, which is then jointly processed by self-attention, enabling deep and bidirectional interactions between all tokens. However, despite its advantages, directly employing this paradigm with Transformers is computationally prohibitive for image fusion due to the quadratic complexity of self-attention. Moreover, its direct application does not guarantee effective cross-modal interaction and integration in image fusion.

In this paper, we propose MMMamba, a novel cross-modal in-context fusion framework for pan-sharpening. Built upon the Mamba architecture, our design achieves linear computational complexity while maintaining strong cross-modal interaction capacity. 
To fully unleash the potential of in-context conditioning within our framework for pan-sharpening task, we introduce a specially designed multimodal interleaved (MI) scanning mechanism that facilitates effective information exchange between the PAN and MS modalities. This method arranges the input sequence so that corresponding PAN and MS tokens are spatially adjacent and can be scanned from different directions. A key advantage of this powerful and unified design is zero-shot task generalization: trained solely on pan-sharpening, MMMamba can perform MS image super-resolution by simply dropping the input PAN modality, without requiring retraining or fine-tuning. Extensive experiments across multiple benchmarks demonstrate that MMMamba consistently outperforms existing state-of-the-art (SOTA) methods both visually and quantitatively.

To summarize, this work offers the following key contributions:

\begin{itemize}
    \item We propose MMMamba, a novel cross-modal in-context fusion framework for pan-sharpening. Built upon the Mamba architecture, it achieves linear complexity and enables bidirectional information flow, while also supporting zero-shot generalization to image super-resolution task.

    \item We are the first to explore the in-context conditioning paradigm in pan-sharpening, enabling deep and efficient cross-modal interactions among all tokens, thereby achieving superior multimodal image fusion results.
    
    \item We design a novel multimodal interleaved (MI) scanning mechanism that facilitates bidirectional information exchange by effectively exploiting complementary cues between PAN and MS modalities.
    
    % We propose MMMamba, a novel cross-modal in-context conditioning framework for pan-sharpening, which can flexibly handle image super-resolution and image colorization tasks in a zero-shot manner.
    
    % \item By introducing the in-context fusion strategy with separate weights, MMMamba ensures linear complexity while enabling efficient bidirectional cross-modal information flow and more thorough feature fusion.
    
    \item Extensive experiments conducted on multiple benchmarks demonstrate that MMMamba consistently outperforms existing SOTA methods across various tasks.
\end{itemize}

\begin{figure*}[ht]
\centering
\includegraphics[width=\textwidth]{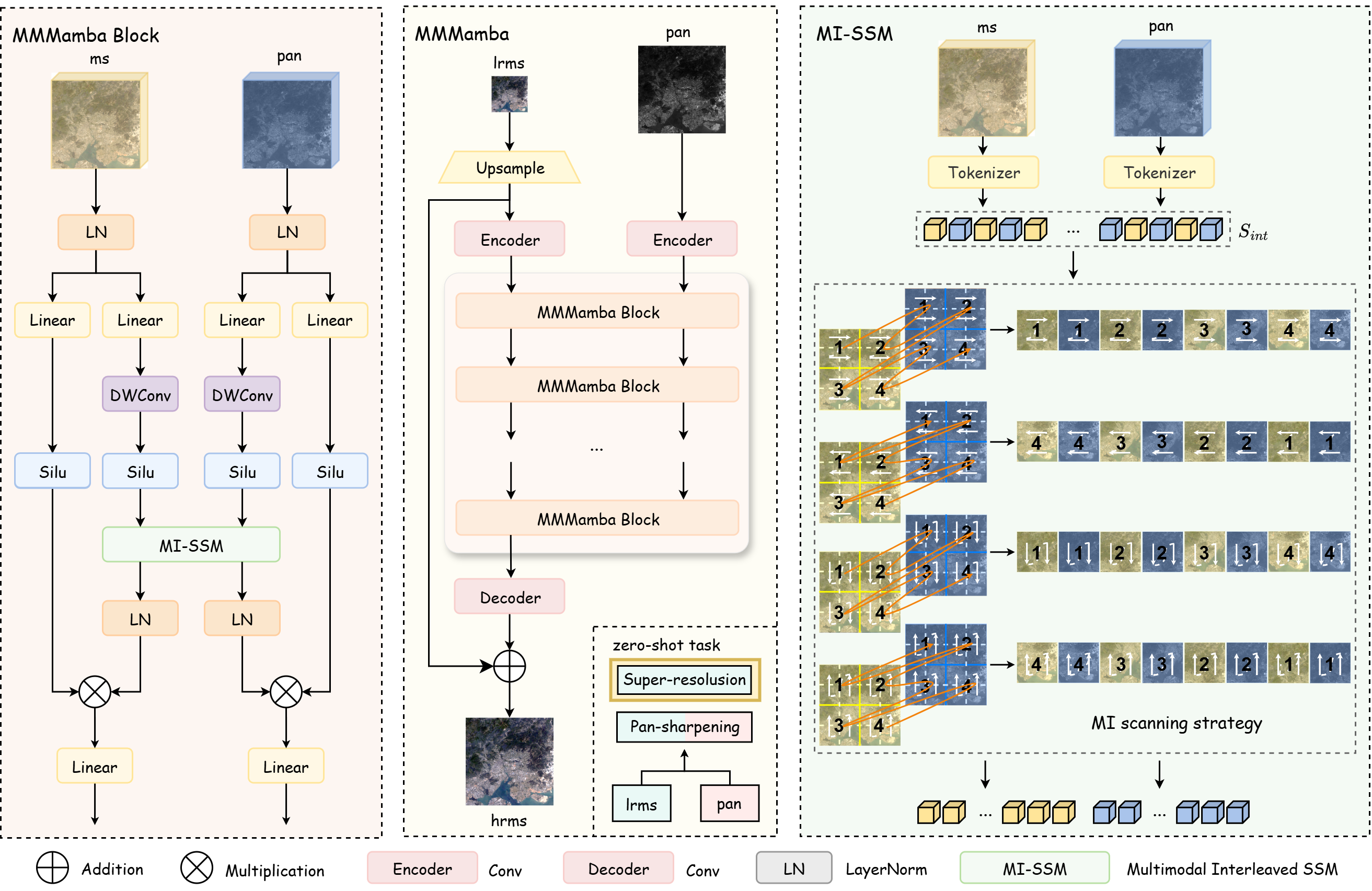}
\caption{\label{framework} The overall framework of our proposed MMMamba, the first exploration of in-context conditioning paradigm in pan-sharpening. This framework enables bidirectional information flow between PAN and MS modalities and supports zero-shot generalization to task like image super-resolution. The proposed MI scanning strategy captures complementary information and facilitates effective cross-modal interaction.}
\end{figure*}

\section{Related Work}

\subsection{Pan-Sharpening}
Pan-sharpening can be categorized into conventional and deep learning-based approaches. Early studies predominantly relied on prior knowledge and handcrafted features, including Component Substitution (CS) \cite{kwarteng1989extracting, gillespie1987color}, Multi-Resolution Analysis (MRA) \cite{schowengerdt1980reconstruction, nunez2002multiresolution}, and Variational Optimization (VO) \cite{fasbender2008bayesian, ballester2006variational}. While traditional approaches offered interpretability and computational efficiency, their limited capacity to model the complex and nonlinear correlations between PAN and MS modalities hindered their performance. The advent of deep learning has reshaped the landscape of pan-sharpening \cite{huang2023dp, zhang2024frequency, meng2025accelerated}. PNN \cite{masi2016pansharpening} first introduced a simple three-layer CNN that achieved promising results. This was followed by a surge of CNN-based sophisticated models, such as PanNet \cite{yang2017pannet}, HFEAN \cite{wang2023learning}, BiMPan \cite{hou2023bidomain}, PIF-Net \cite{meng2024progressive}. More recently, the integration of Transformer-based models, such as Panformer \cite{zhou2022panformer}, CMINet \cite{wang2024cross},  LFormer \cite{hou2024linearly}, and FCSA \cite{wu2025fully}, introduced self-attention mechanisms to capture long-range dependencies, significantly improving the modeling of spatial relationships. 

\subsection{State Space Model}
State Space Models (SSMs) have recently emerged as a powerful alternative to CNNs and Transformers, owing to their long-range dependencies with linear computational complexity. S4 \cite{gu2021efficiently} introduced diagonal state-space parameterizations for efficient parallelization, and Mamba \cite{gu2023mamba} further incorporated a dynamic selection mechanism to enhance training and sequence modeling. Recent research has successfully adapted SSMs to the visual domain by reshaping images into sequential representations and integrating specialized scanning mechanisms. Specifically, Vmamba \cite{liu2024vmamba} and Vision Mamba \cite{zhu2024vision} employed directionally-aware scanning schemes to effectively model spatial structures, facilitating the integration of contextual information from various perspectives. LEVM \cite{cao2024novel} introduced a local-enhanced vision Mamba block tailored for image fusion tasks, which strengthened local spatial perception and improved the integration of spatial and spectral information. Pan-Mamba \cite{he2025pan} is the pioneering work that introduces Mamba into pan-sharpening, effectively modeling long-range dependencies and cross-modal correlations for efficient global processing and superior spectral–spatial fusion. These approaches, although effective, are typically limited to a single task, such as image fusion or super-resolution, and cannot flexibly handle zero-shot generalization to other tasks. Moreover, the scanning strategies employed in these methods fail to facilitate efficient cross-modal information exchange, thereby constraining the quality of the fusion results.

\section{Methodology}

\subsection{Problem Formulation}
Pan-sharpening seeks to fuse the complementary information between the multispectral (MS) image $I_{lms} \in \mathbb{R}^{H / s \times W / s \times C}$ and the panchromatic (PAN) image $I_{p} \in \mathbb{R}^{H \times W \times 1}$ in order to produce the high-resolution multispectral (HRMS) image $I_{hms} \in \mathbb{R}^{H \times W \times C}$. Here, $H$, $W$, and $C$ represent the image height, width, and number of spectral channels, respectively, and $s$ defines the spatial resolution ratio between $I_{lms}$ and $I_{hms}$, which is set to 4. The overall architecture of MMMamba is shown in Figure \ref{framework}.

\subsection{Network Architecture}
Given the upsampled LRMS image $I_{ms} \in \mathbb R^{H \times W \times C}$ and PAN image $I_{p} \in \mathbb R^{H \times W \times 1}$, both inputs are first passed through separate gated convolutional encoders \cite{hornet}, denoted as $E_{\varphi}^{ms}$ and $E_{\varphi}^p$, to extract shallow features from their respective modalities, resulting in $F_{ms} \in \mathbb{R}^{B \times C \times H \times W}$ and $F_p \in \mathbb{R}^{B \times C \times H \times W}$:
\begin{equation}
F_{ms}=E_{\varphi}^{ms}\left(I_{ms}\right),
\end{equation}
\begin{equation}
F_p=E_{\varphi}^p\left(I_{p}\right).
\end{equation}

\subsubsection{MMMamba Blocks}
The shallow features $F_{ms}$ and $F_p$, derived from the MS and PAN modalities, are then independently processed through a series of MMMamba blocks, which enable deep cross-modal interaction and efficient in-context fusion.

Specifically, $F_{ms}$ and $F_p$ first undergo layer normalization, followed by a linear projection to transform the feature dimensions. The outputs are denoted as $F_{ms}^{ln} \in \mathbb R^{B \times H \times W \times C}$ and $F_{p}^{ln} \in \mathbb R^{B \times H \times W \times C}$:
\begin{equation}
F_{ms}^{ln} = \text{Linear} \left( \text{LN}(F_{ms}) \right),
\end{equation}
\begin{equation}
F_{p}^{ln} = \text{Linear} \left( \text{LN}(F_{p}) \right).
\end{equation}

Next, these normalized and projected features are processed by depth-wise convolutional layers (DWConv), and then activated using the sigmoid linear unit (SiLU) function, yielding $F_{ms}^{silu} \in \mathbb R^{B \times C \times H \times W}$ and $F_{p}^{silu} \in \mathbb R^{B \times C \times H \times W}$:
\begin{equation}
F_{ms}^{silu} = \text{SiLU} \left( \text{DWConv}(F_{ms}^{ln}) \right),
\end{equation}
\begin{equation}
F_{p}^{silu} = \text{SiLU} \left( \text{DWConv}(F_{p}^{ln}) \right).
\end{equation}

\subsubsection{Multimodal Interleaved (MI) SSM}
The multimodal interleaved scanning operation, denoted as $\operatorname{MI-Scan}(\cdot)$, is applied to enable effective cross-modal information exchange and to capture complementary characteristics between the MS and PAN modalities. The details of MI-SSM are illustrated in the right part of Figure~\ref{framework}.

\noindent \textbf{Tokenization\ }
Initially, the features $F_{ms}^{silu}$ and $F_{p}^{silu}$ from the MS and PAN modalities are tokenized into non-overlapping patches. These patches are then interleaved in four predefined directions: ``left-to-right and up-to-down" (``\text{ltr\_utd}"), ``up-to-down and left-to-right" (``\text{utd\_ltr}"), ``right-to-left and down-to-up" (``\text{rtl\_dtu}"), and ``down-to-up and right-to-left" (``\text{dtu\_rtl}"):
\begin{align}
T_{ms}^{\text{ltr\_utd}},\ T_{p}^{\text{ltr\_utd}} &= \operatorname{Tokenize}\left( F_{ms}^{silu},\ F_{p}^{silu} \right)_{\text{ltr\_utd}}, \\
T_{ms}^{\text{utd\_ltr}},\ T_{p}^{\text{utd\_ltr}} &= \operatorname{Tokenize}\left( F_{ms}^{silu},\ F_{p}^{silu} \right)_{\text{utd\_ltr}}, \\
T_{ms}^{\text{rtl\_dtu}},\ T_{p}^{\text{rtl\_dtu}} &= \operatorname{Tokenize}\left( F_{ms}^{silu},\ F_{p}^{silu} \right)_{\text{rtl\_dtu}}, \\
T_{ms}^{\text{dtu\_rtl}},\ T_{p}^{\text{dtu\_rtl}} &= \operatorname{Tokenize}\left( F_{ms}^{silu},\ F_{p}^{silu} \right)_{\text{dtu\_rtl}},
\end{align}
% where $S_{p1}$, $S_{p2} \in \mathbb{R}^{B \times C \times H_g \times W_g \times s \times s}$:
% \begin{align}
% S_{p1}, S_{p2} = \text{Reshape}(S_{int1}, S_{int2}).
% \end{align}
where $T_{ms}^{k}, T_{p}^{k} \in \mathbb{R}^{B \times C \times H_g \times W_g \times s \times s}$, and $k \in \{ \text{ltr\_utd}, \text{utd\_ltr}, \text{rtl\_dtu}, \text{dtu\_rtl} \}$. Here, $H_g$ and $W_g$ denote the number of rows and columns in the patch grid, respectively, with $H_g = H / s$ and $W_g = W / s$, and $s$ represents the spatial size of each patch.

For each direction, patch-wise interleaving is performed to form a fused sequence of patches from both the MS and PAN modalities:
\begin{equation}
{S}_{int}^{\text{ltr\_utd}} = \operatorname{Interleave}\left( T_{ms}^{\text{ltr\_utd}}, T_{p}^{\text{ltr\_utd}} \right),
\end{equation}
\begin{equation}
{S}_{int}^{\text{utd\_ltr}} = \operatorname{Interleave}\left( T_{ms}^{\text{utd\_ltr}}, T_{p}^{\text{utd\_ltr}} \right),
\end{equation}
\begin{equation}
{S}_{int}^{\text{rtl\_dtu}} = \operatorname{Interleave}\left( T_{ms}^{\text{rtl\_dtu}}, T_{p}^{\text{rtl\_dtu}} \right),
\end{equation}
\begin{equation}
{S}_{int}^{\text{dtu\_rtl}} = \operatorname{Interleave}\left( T_{ms}^{\text{dtu\_rtl}}, T_{p}^{\text{dtu\_rtl}} \right).
\end{equation}

The sequences from all four directions are then concatenated to generate the interleaved sequence:
\begin{equation}
{S}_{int} = \operatorname{Concat}\left( {S}_{int}^{\text{ltr\_utd}}, {S}_{int}^{\text{utd\_ltr}}, {S}_{int}^{\text{rtl\_dtu}}, {S}_{int}^{\text{dtu\_rtl}} \right),
\end{equation}
where $S_{{int}} \in \mathbb{R}^{B \times 4 \times C \times L}$, with $L = 2\times H \times W$.

\noindent \textbf{MI Scanning Strategy\ }
The MI scanning strategy first splits the $S_{int}$ into sequence of four directions ${S}_{int}^{\text{ltr\_utd}}, {S}_{int}^{\text{utd\_ltr}}, {S}_{int}^{\text{rtl\_dtu}}, {S}_{int}^{\text{dtu\_rtl}}$. Each sequence is reshaped into $\mathbb{R}^{B \times C \times H_g \times W_g \times 2 \times s^2 }$:
\begin{equation}
{S}_{int}^{\text{ltr\_utd}}, {S}_{int}^{\text{utd\_ltr}}, {S}_{int}^{\text{rtl\_dtu}}, {S}_{int}^{\text{dtu\_rtl}} = \operatorname{Split}(S_{int}).
\end{equation}

The sequences are then split into two parts to perform cross-modal MI scanning:
\begin{equation}
{S}_{int\_1}^{\text{ltr\_utd}}, {S}_{int\_2}^{\text{ltr\_utd}} = \operatorname{Split}({S}_{int}^{\text{ltr\_utd}}),
\end{equation}
\begin{equation}
{S}_{int\_1}^{\text{utd\_ltr}}, {S}_{int\_2}^{\text{utd\_ltr}} = \operatorname{Split}({S}_{int}^{\text{utd\_ltr}}),
\end{equation}
\begin{equation}
{S}_{int\_1}^{\text{rtl\_dtu}}, {S}_{int\_2}^{\text{rtl\_dtu}} = \operatorname{Split}({S}_{int}^{\text{rtl\_dtu}}),
\end{equation}
\begin{equation}
{S}_{int\_1}^{\text{dtu\_rtl}}, {S}_{int\_2}^{\text{dtu\_rtl}} = \operatorname{Split}({S}_{int}^{\text{dtu\_rtl}}),
\end{equation}
where ${S}_{int\_1}^{k}, {S}_{int\_2}^{k} \in \mathbb{R}^{B \times C \times H_g \times W_g \times s \times s}$, and $k \in \{ \text{ltr\_utd}, \text{utd\_ltr}, \text{rtl\_dtu}, \text{dtu\_rtl} \}$.

Next, these sequences are divided into multiple local windows. For each local window, selective scanning is first applied to ${S}_{int\_1}^{\text{ltr\_utd}}$ using the ``\text{ltr\_utd}" scanning direction. After completing this, the scanning is transferred to the corresponding local window of the ${S}_{int\_2}^{\text{ltr\_utd}}$, where the same selective scanning is executed. Once finished, the process returns to the next local window of ${S}_{int\_1}^{\text{ltr\_utd}}$ and repeats the same procedure. This alternating scanning continues for all local windows:
\begin{align}
S_{mi1}^{\text{ltr\_utd}}, S_{mi2}^{\text{ltr\_utd}} = \operatorname{MI-Scan}({S}_{int\_1}^{\text{ltr\_utd}}, {S}_{int\_2}^{\text{ltr\_utd}}),
\end{align}
where $S_{mi1}^{\text{ltr\_utd}},\ S_{mi2}^{\text{ltr\_utd}} \in \mathbb{R}^{B \times C \times L'}$, and $L' = H \times W$.

The scanning strategy then proceeds with three additional directions, ``\text{utd\_ltr}", ``\text{rtl\_dtu}", and ``\text{dtu\_rtl}". Such multi-directional scanning approach enhances cross-modal interaction and enables better exploitation of complementary information:
\begin{equation}
S_{mi1}^{\text{utd\_ltr}}, S_{mi2}^{\text{utd\_ltr}} = \operatorname{MI-Scan}({S}_{int\_1}^{\text{utd\_ltr}}, {S}_{int\_2}^{\text{utd\_ltr}}),
\end{equation}
\begin{equation}
S_{mi1}^{\text{rtl\_dtu}}, S_{mi2}^{\text{rtl\_dtu}} = \operatorname{MI-Scan}({S}_{int\_1}^{\text{rtl\_dtu}}, {S}_{int\_2}^{\text{rtl\_dtu}}),
\end{equation}
\begin{equation}
S_{mi1}^{\text{dtu\_rtl}}, S_{mi2}^{\text{dtu\_rtl}} = \operatorname{MI-Scan}({S}_{int\_1}^{\text{dtu\_rtl}}, {S}_{int\_2}^{\text{dtu\_rtl}}).
\end{equation}

The outputs of the MI-SSM are computed by summing the results of the four directional scans:
\begin{align}
S_{mi1}^{\text{out}} &= S_{mi1}^{\text{ltr\_utd}} + S_{mi1}^{\text{utd\_ltr}} + S_{mi1}^{\text{rtl\_dtu}} + S_{mi1}^{\text{dtu\_rtl}}, \\
S_{mi2}^{\text{out}} &= S_{mi2}^{\text{ltr\_utd}} + S_{mi2}^{\text{utd\_ltr}} + S_{mi2}^{\text{rtl\_dtu}} + S_{mi2}^{\text{dtu\_rtl}},
\end{align}
where $S_{mi1}^{\text{out}},\ S_{mi2}^{\text{out}} \in \mathbb{R}^{B \times C \times L}$.

The output features from the MI-SSM, $S_{mi1}^{\text{out}}$ and $S_{mi2}^{\text{out}}$, are subsequently combined with the SiLU-activated projections of the normalized $F_{ms}^{ln}$ and $F_{p}^{ln}$ respectively through element-wise multiplication and summation:
\begin{equation}
F_{ms}^{mm} = \text{LN}(S_{mi1}^{\text{out}}) \odot \text{SiLU}(F_{ms}^{ln}),
\end{equation}
\begin{equation}
F_{p}^{mm} = \text{LN}(S_{mi2}^{\text{out}}) \odot \text{SiLU}(F_{p}^{ln}),
\end{equation}
where $F_{ms}^{mm},\ F_{p}^{mm} \in \mathbb{R}^{B \times C \times L}$.

These features are then passed through linear projections and reshaped to produce $F_{ms}^{out},\ F_{p}^{out} \in \mathbb{R}^{B \times C \times H \times W}$, delivering the final output of the MMMamba block:
\begin{align}
F_{ms}^{\text{out}},\ F_{p}^{\text{out}} = \text{Linear}(F_{ms}^{mm}),\ \text{Linear}(F_{p}^{mm}).
\end{align}

The resulting outputs are forwarded to the subsequent MMMamba block, which progressively refines the multimodal representations and enriches cross-modal feature interactions, effectively exploiting complementary cues between modalities and enabling efficient in-context fusion.

Afterward, a convolutional decoder $D_{\varphi}$ is applied to the output of the last MMMamba block to generate the final MS feature $F_{ms}^{\text{final}}$:
\begin{equation}
F_{ms}^{\text{final}}=D_{\varphi}\left(F_{ms}^{\text{out\_last}}\right),
\end{equation}
where $F_{ms}^{\text{out\_last}}$ denotes the output of the last MMMamba block.

Finally, the HRMS result is obtained by adding $F_{ms}^{\text{final}}$ to the upsampled LRMS image $I_{ms} \in \mathbb{R}^{H \times W \times C}$:
\begin{equation}
I_{hms} = F_{ms}^{\text{final}} + I_{ms}.
\end{equation}

\subsection{Loss Function}
We employ the $\mathcal{L}_{1}$ as the loss function \cite{zhao2016loss}. The predicted HRMS image is denoted by $I_{hms}$, and the corresponding ground truth is defined by $I_{gt}$. The loss can be expressed as:
\begin{equation}
\mathcal{L} = \| I_{gt} - I_{hms} \|_1. 
\end{equation}

\begin{table*}[ht]
\centering
\resizebox{\textwidth}{!}{%
\begin{tblr}{
  colspec = {lcccccccccccc}, % 第一列左对齐，其他列居中
  cell{1}{1} = {r=2}{},
  cell{1}{2} = {c=4}{c},
  cell{1}{6} = {c=4}{c},
  cell{1}{10} = {c=4}{c},
  vline{2-3,5-6,7,10} = {1, 2-5}{0.08em},
  vline{3-13} = {2-16}{0.08em},
  vline{2-13} = {3-16}{0.08em},
  hline{1,2-3} = {-}{0.08em},
  hline{2-3} = {2-13}{0.05em},
  hline{8,15,16} = {-}{0.05em},
}
\textbf{Methods}    &\textbf{ WorldView-II}     &                 &                 &                 & \textbf{GaoFen2}          &                 &                 &                 & \textbf{Worldview-III}    &                 &                 &                 \\
           & PSNR$\uparrow$             & SSIM$\uparrow$            & SAM$\downarrow$             & ERGAS$\downarrow$           & PSNR$\uparrow$             & SSIM$\uparrow$            & SAM$\downarrow$             & ERGAS$\downarrow$           & PSNR$\uparrow$             & SSIM$\uparrow$            & SAM$\downarrow$             & ERGAS$\downarrow$           \\
IHS \cite{haydn1982application}       & 35.2962          & 0.9027          & 0.0461          & 2.0278          & 38.1754          & 0.9100          & 0.0243          & 1.5336          & 22.5579          & 0.5354          & 0.1266          & 8.3616          \\
Brovey \cite{gillespie1987color}    & 35.8646          & 0.9216          & 0.0403          & 1.8238          & 37.7974          & 0.9026          & 0.0218          & 1.3720          & 22.5060          & 0.5466          & 0.1159          & 8.2331          \\
SFIM \cite{liu2000smoothing}      & 34.1297          & 0.8975          & 0.0439          & 2.3449          & 36.9060          & 0.8882          & 0.0318          & 1.7398          & 21.8212          & 0.5457          & 0.1208          & 8.9730          \\
GFPCA  \cite{liao2015two}    & 34.5581          & 0.9038          & 0.0488          & 2.1411          & 37.9443          & 0.9204          & 0.0314          & 1.5604          & 22.3344          & 0.4826          & 0.1294          & 8.3964          \\
LRTCFPan \cite{wu2023lrtcfpan}   & 34.7756          & 0.9112          & 0.0426          & 2.0075          & 36.9253          & 0.8946           & 0.0332          & 1.7060          & 22.1574           & 0.5735          & 0.1380          & 8.6796           \\
SRPPNN \cite{cai2020super}    & 41.4538          & 0.9679          & 0.0233          & 0.9899          & 47.1998          & 0.9877          & 0.0106          & 0.5586          & 30.4346          & 0.9202          & 0.0770          & 3.1553          \\
INNformer \cite{zhou2022pan}  & 41.6903          & 0.9704          & 0.0227          & 0.9514          & 47.3528          & 0.9893          & 0.0102          & 0.5479          & 30.5365          & 0.9225          & 0.0747          & 3.0997          \\

FAME \cite{he2024frequency}     & 42.0262          & 0.9723          & 0.0215          & 0.9172          & \underline{47.6721}          & \underline{0.9898}          & {0.0098}           & \underline{0.5242}          & 30.9903          & 0.9287          & \underline{0.0697}          & 2.9531          \\

WaveletNet \cite{WaveletNet}  & 41.9131          & 0.9715          & 0.0220          & 0.9274          & \underline{47.5907}          & \underline{0.9894}          & {0.0099}           & \underline{0.5310}          & 30.9139          & 0.9279          & \underline{0.0710}          & 2.9770          \\

SFINet++ \cite{zhou2024general}     & 41.8115          & 0.9731          & 0.0220          & 0.9489          & \underline{47.5344}          & \underline{0.9906}          & {0.0100}           & \underline{0.5356}          & 30.7665          & 0.9261          & \underline{0.0732}          & 3.0217          \\

Pan-Mamba \cite{he2025pan}   & \underline{42.2354}          & \underline{0.9729}          & \underline{0.0212}          & \underline{0.8975}          & 47.6453          & 0.9894          & 0.0103          & 0.5286          & \underline{31.1740}          & \underline{0.9302}          & 0.0698          & \underline{2.8910}          \\
% CDFCNet \cite{li2025pan}   & \underline{42.2406}          & \underline{0.9733}          & \underline{0.0209}          & \underline{0.8918}          & 47.8423          & 0.9902          & \textbf{0.0097}          & 0.5134          & \underline{31.2386}          & \underline{0.9323}          & 0.0694          & \underline{2.8651}          \\
CFLIHPs \cite{wang2025towards}   & \underline{41.9077}          & \underline{0.9712}          & \underline{0.0220}          & \underline{0.9284}          & 47.3824          & 0.9892          & 0.0102          & 0.5409          & \underline{30.8341}          & \underline{0.9269}          & 0.0737          & \underline{2.9980}          \\
Ours       & \textbf{42.3120} & \textbf{0.9733} & \textbf{0.0209} & \textbf{0.8888} & \textbf{47.9932} & \textbf{0.9902} & \textbf{0.0098} & \textbf{0.5126} & \textbf{31.2311} & \textbf{0.9305}          & \textbf{0.0687} & \textbf{2.8950} \\
\end{tblr}%
}
\caption{Quantitative comparison on three datasets. The best results are highlighted in \textbf{bold}. $\uparrow$ signifies better performance with larger values, while $\downarrow$ indicates improved performance with smaller values.}
\label{tab:1}
\end{table*}

\begin{table*}[ht]
\centering
\resizebox{\textwidth}{!}{%
\begin{tblr}{
  cells = {c},
  vline{2} = {-}{0.05em},
  hline{1,5} = {-}{0.08em},
  hline{2} = {-}{0.05em},
}
\textbf{Metrics} & \textbf{IHS}   & \textbf{Brovey} &  \textbf{SFIM}   & \textbf{GFPCA}  &

\textbf{LRTCFPan}  & \textbf{SRPPNN} & \textbf{INNformer}  & \textbf{FAME} & \textbf{WaveletNet} & \textbf{SFINet++}  & \textbf{Pan-Mamba} & \textbf{CFLIHPs}  & \textbf{Ours}            \\
$D_{\lambda}$$\downarrow$     & 0.0770 & 0.1378 & 0.0822 & 0.0914 &
 0.1170 & 0.0767 & 0.0782 & 0.0674    & 0.0700  & 0.0673   & \underline{0.0652}  & {0.0678} & \textbf{0.0656} \\
$D_S
$$\downarrow$     & 0.2985 & 0.2605 & 0.1121& 0.1635 & 0.2024 & 0.1162 & 0.1253 & 0.1121    & \underline{0.1063} & {0.1108}  & 0.1129 & {0.1170} & \textbf{0.1113} \\
QNR$\uparrow$    & 0.6485 & 0.6390 & 0.8214 & 0.7615 & 0.7063  & 0.8173 & 0.8073 & 0.8291    & \underline{0.8327}  & \underline{0.8471}  & 0.8306 & {0.8287}  & \textbf{0.8312} 
\end{tblr}
}
\caption{Evaluation of our method on real-world full-resolution scenes from the GF2 dataset.}
\label{tab:2}
\end{table*}

\begin{figure*}[!t]
    \centering
    \includegraphics[width=16cm]{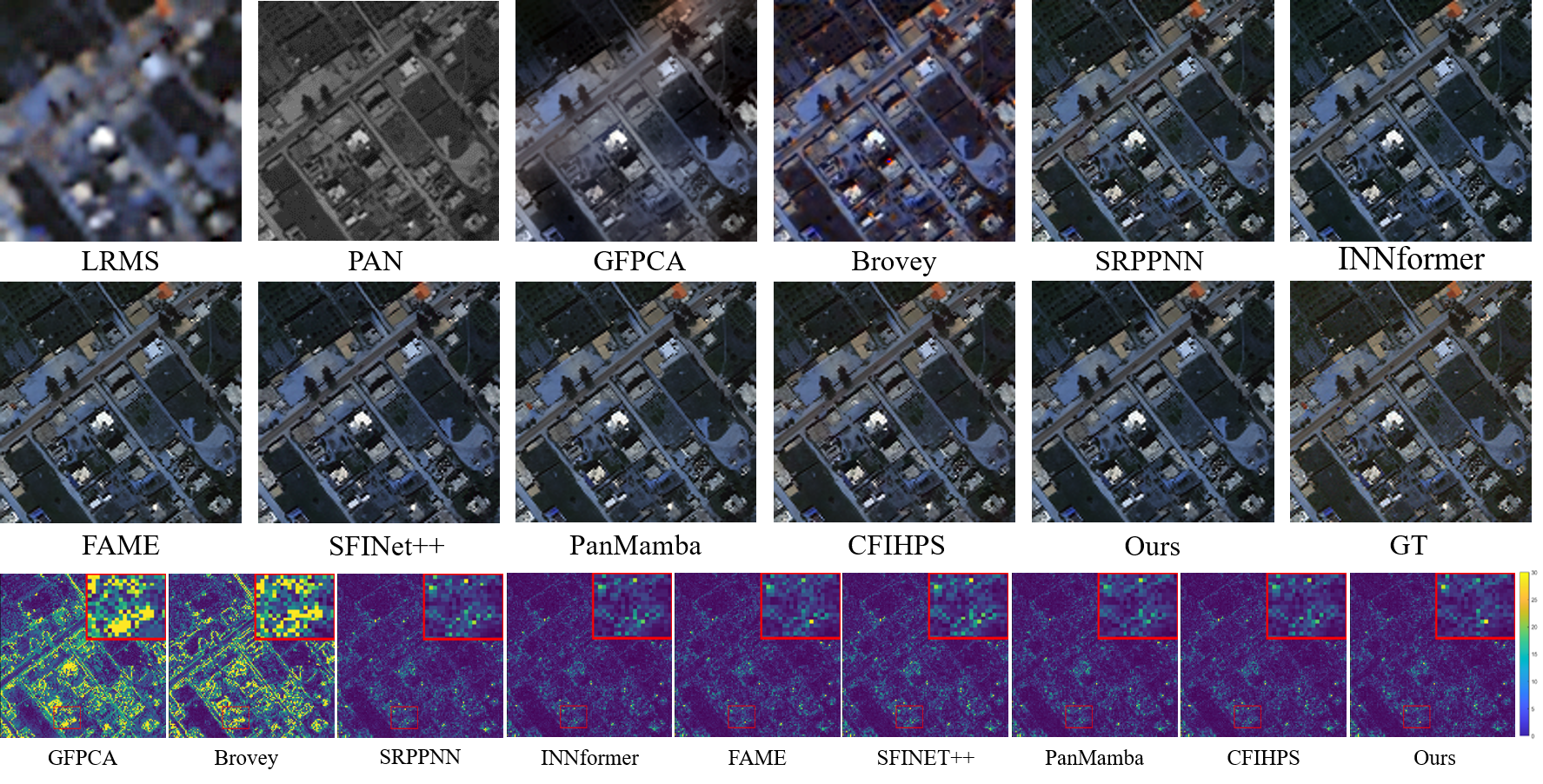}
    \caption{Visual comparison of all methods on WV3. The last row visualizes the MSE residues between the pan-sharpening results and the ground truth. }
    \label{wv3_result}
\end{figure*}

\begin{figure}[h]
  \centering
  \includegraphics[width=\linewidth]{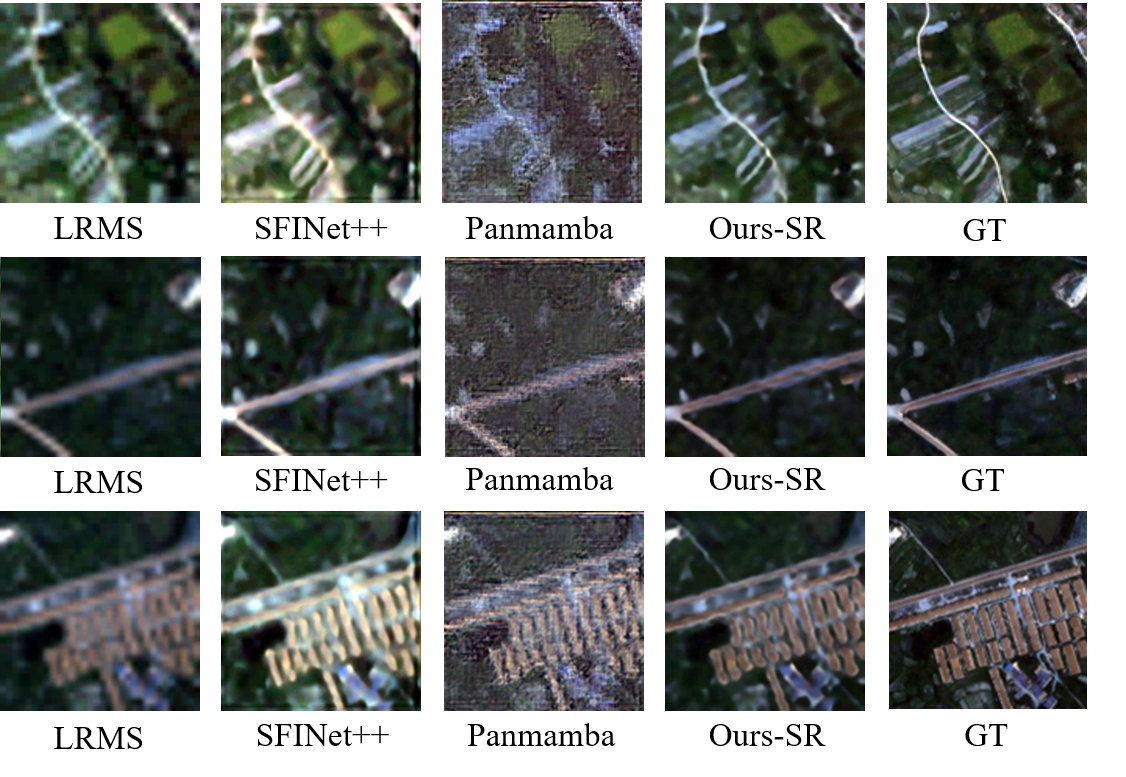}
  \caption{The visual comparison of the zero-shot image super-resolution results on the WV2 dataset.}
  \label{zero-shot}
\end{figure}

\section{Experiments}
\begin{table*}[t!]
\centering
\resizebox{0.85\textwidth}{!}{% Using resizebox to ensure the table fits within a single column
\begin{tabular}{@{}llcccccc@{}}
\toprule
\multirow{2}{*}{\textbf{ID}} & \multirow{2}{*}{\textbf{Methods / Variant}} & \multicolumn{4}{c}{\textbf{Performance Metrics}} & \multicolumn{2}{c}{\textbf{Efficiency Metrics}} \\ 
\cmidrule(l{2pt}r{2pt}){3-6} \cmidrule(l{2pt}r{2pt}){7-8}
 &  & PSNR $\uparrow$ & SSIM $\uparrow$ & SAM $\downarrow$ & ERGAS $\downarrow$ & Params (M) $\downarrow$ & FLOPs (G) $\downarrow$ \\ \midrule
\multicolumn{8}{l}{\textit{Ablation on Core Paradigm \& Backbone}} \\ \midrule
M0 & \textbf{MMMamba (Full Model / Baseline)} & \textbf{42.3120} & \textbf{0.9733} & \textbf{0.0209} & \textbf{0.8888} & 0.2453 & 5.0616 \\
M1 & w/o Mamba (use Transformer) & 41.3995 & 0.9675 & 0.0235 & 0.9862 & \textbf{0.3684} & \textbf{3.9206} \\
M2 & w/o In-context Fusion (use Channel Concat.) & 41.2898 & 0.9672 & 0.0237 & 0.9976 &0.2704 & 5.0275 \\
M3 & w/o Interleaving (use Sequential Concat.) & 36.4702 & 0.9107 & 0.0302 & 1.5550 & 0.2432&5.0275  \\ \midrule
\multicolumn{8}{l}{\textit{Ablation on Scanning Strategy}} \\ \midrule
M4 & w/o Multi-direction (use 1-way Scan) & 42.0965 & 0.9723 & 0.0214 & 0.0989 & 0.2432 & \textbf{5.0275} \\
M5 & w/o Local Scan (use Global Scan) & 42.1998 & 0.9729 & 0.0211 & 0.8972 & 0.2432& 5.0277 \\
\bottomrule
\end{tabular}%
}
\caption{Ablation study of the MMMamba model on the WV2 dataset. `$\uparrow$’ indicates that higher is better, while `$\downarrow$’ indicates that lower is better. \textbf{Bold} marks the best result in each column. All models are trained and evaluated under identical settings.}
\label{tab:ablation_english}
\end{table*}

\subsection{Datasets and Benchmark}
We conducted experiments using data from three satellites: WorldView-II (WV2), GaoFen2 (GF2), and WorldView-III (WV3). These datasets provide a variety of resolutions and scenes, including industrial areas and natural landscapes from WV2, mountains and rivers from GF2, and urban environments from WV3. As ground truth was not available, we generated all test datasets at a reduced resolution according to the Wald protocol. We compared our proposed model against several traditional methods, specifically GFPCA \cite{liao2015two}, LRTCFPan \cite{wu2023lrtcfpan}, Brovey \cite{gillespie1987color}, IHS \cite{haydn1982application}, and SFIM \cite{liu2000smoothing}, as well as recent deep learning-based methods, including SRPPNN \cite{cai2020super}, INNformer \cite{zhou2022pan}, FAME \cite{he2024frequency}, SFINet++ \cite{zhou2024general}, WaveletNet \cite{WaveletNet}, Pan-Mamba \cite{he2025pan}, and CFLIHPs \cite{wang2025towards}.
The performance was quantitatively evaluated using a combination of full-reference and no-reference metrics. The full-reference metrics were Peak Signal-to-Noise Ratio (PSNR), Structural Similarity Index (SSIM), Spectral Angle Mapper (SAM), and the relative dimensionless global error in synthesis (ERGAS). The no-reference metrics were the spatial distortion index ($D_S$), the spectral distortion index ($D_\lambda$), and the Quality with No Reference ($QNR$) index.

\subsection{Implement Details}
We implemented the model in PyTorch and conducted all training on a single Nvidia V100 GPU. For optimization, we used the Adam optimizer with a gradient clipping norm of 4.0 to ensure training stability. The learning rate was initialized to $5 \times 10^{-4}$ and adjusted using a cosine decay schedule, which reduced it to $5 \times 10^{-8}$ by the final epoch. To account for variations in data volume, we trained the model for 200 epochs on the WorldView-II dataset and 500 epochs on both the GaoFen2 and WorldView-III datasets.

\subsection{Comparison With SOTA Methods}
\subsubsection{Evaluation on Reduced-Resolution Scene}
Table \ref{tab:1} summarizes the quantitative results of MMMamba in comparison with existing methods across three benchmark datasets, demonstrating its superior performance over existing SOTA techniques across multiple evaluation metrics. In particular, our approach achieves notable gains in PSNR, outperforming the CFLIHPs by 0.40 dB and 0.61 dB on the WV2 and GF2 datasets, respectively. Figure \ref{wv3_result} presents qualitative results from the WV3 dataset. The residual plots produced by our method exhibit the lowest brightness, reflecting a high degree of consistency with the ground truth. Additionally, our approach yields sharper edges and more accurate spectral details, further emphasizing its advantage over competing methods.

\subsubsection{Evaluation on Full-Resolution Scene}
We further conducted a full-resolution evaluation under real-world conditions to assess the generalization capability of our method. This experiment was carried out on the full-resolution GF2 (FGF2) datasets, where no-reference quality metrics were employed due to the absence of ground truth references. The FGF2 dataset was utilized in its original form without any downsampling, providing a testing environment that closely replicates real-world image degradation. As summarized in Table~\ref{tab:2}, our method consistently outperforms other approaches across all three metrics, demonstrating its strong generalization performance in real-world scenarios.

\subsection{Zero-Shot Task Generalization}
To evaluate MMMamba's zero-shot generalization capabilities, we tested it on MS image super-resolution. Although trained exclusively on pan-sharpening, MMMamba can perform this tasks without any retraining or fine-tuning. By leveraging its in-context fusion mechanism, it adapts by simply omitting one input modality—performing super-resolution when given only the MS image.

We also compared our approach with other deep learning models. Since these models cannot inherently work with a single input, we had to adapt them. For the super-resolution task, we fed the MS image into both the PAN and MS encoders during inference. 

The qualitative results for the zero-shot super-resolution task are presented in Figure \ref{zero-shot}. As illustrated in the figure, our proposed method generates visually compelling results, successfully reconstructing finer details and sharper edges. In contrast, the outcomes from the adapted SFINet++ and Pan-Mamba methods appear comparatively blurry, with a noticeable loss of textural information.

The quantitative metrics, summarized in Table \ref{zero-shot_table}, provide further evidence of our model's effectiveness. Our approach consistently outperforms the compared methods across all evaluation criteria. Notably, our model achieves a PSNR of 36.49 dB and an SSIM of 0.9114. These scores not only surpass those of the other deep learning-based methods, SFINet++ and Pan-Mamba, but also exceed the performance of the traditional Bicubic interpolation. Furthermore, our method yields the lowest error values with a SAM of 0.0299 and an ERGAS of 1.5515, indicating superior spectral and radiometric fidelity in the reconstructed images.

% To evaluate its zero-shot generalization capabilities, we apply MMMamba to MS image super-resolution and PAN image colorization. Although trained exclusively on pan-sharpening, MMMamba can perform these tasks without retraining or fine-tuning by simply omitting one input modality, leveraging its in-context fusion mechanism. Specifically, it performs super-resolution when given only the MS image and colorization when given only the PAN image. The visual results are presented in Figure \ref{zero-shot}.  We also compare to other deeplearning methods by replacing there pan feature with ms feature and pan feature with ms feature for super-resolution and colorization tasks, we also retrain their methods by deleting skip connection for zero-shot colorization evaluation. As shown in the Figure, in super-resolution tasks, our methods outperform other methods, while in colorization, other method cannot apply color to pan image, our method, while not accurate, demonstrate the potential of image colorization ability.
% Notably, the metric of super-resolution in some cases even surpassing the traditional pan-sharpening task.

\subsection{Ablation Experiments}
We conducted ablation studies on the WV2 dataset to validate core components, as presented in Table~\ref{tab:ablation_english}.
\subsubsection{Ablation on Core Paradigm}
Our analysis confirms the effectiveness of the core design (M1). We replaced the Mamba operator with a self-attention (SA) based block to compare their effectiveness. For a fair comparison under a similar computational budget, we built a computationally matched SA by incorporating sequence downsampling (via pixel shuffling) and a linear projection before the attention mechanism. 
Replacing the \textbf{Mamba operator} with this self-attention module degrades performance, underscoring Mamba's linear ($O(N)$) efficiency for this task~\cite{he2025pan}. 

The proposed \textbf{in-context fusion} is also crucial: substituting it with naive channel-wise concatenation (M2) prevents effective cross-modal interaction, causing the PSNR to drop to 41.28 dB. 

The necessity of our \textbf{interleaved design} is demonstrated by replacing it with sequential token concatenation (M3), which caused a huge performance decrease. The interleaved approach places tokens from corresponding spatial positions of each modality adjacent to one another, enabling direct and efficient information exchange. Conversely, sequential concatenation separates these corresponding tokens, causing severe information decay as the signal propagates over a long distance within the Mamba state.

\subsubsection{Ablation on Scanning Strategy}
\paragraph{Benefits of Multi-Directional Scanning}
To validate multi-directional scanning, we simplified it to a single, unidirectional scan (M4). The results show a performance drop with negligible change in computational cost. This confirms that aggregating contextual information from multiple directions allows the model to build a more comprehensive and robust feature representation, essential for 2D spatial data.

\paragraph{Effectiveness of the Local Window Scan}
We compared our local window scan against a standard global scan (M5), where the latter showed a slight decline in performance (PSNR drops to 42.19 dB). This experiment demonstrates that our modification successfully introduces a crucial inductive bias of locality into the Mamba operator.

% \begin{table}[h]
% \centering
% \resizebox{0.85\linewidth}{!}{
% \begin{tabular}{lcccccc}
% \hline
% \textbf{Methods} & \textbf{PSNR$\uparrow$}  & \textbf{SSIM$\uparrow$} & \textbf{SAM$\downarrow$} & \textbf{ERGAS$\downarrow$}  \\
% \hline
% Bicubic &34.0869 &0.8726 &0.0397 &2.1202\\
% SFINet++ & 33.3047 & 0.8679 & 0.0439 & 2.3105\\
% Pan-Mamba & 30.5913 & 0.7656 & 0.0524 & 3.1224 \\
% Ours & \textbf{36.4892} & \textbf{0.9114} & \textbf{0.0299} & \textbf{1.5515}\\
% \hline
% \end{tabular}}
% \caption{Comparison results on the WV2 dataset for zero-shot image super-resolution evaluation.}
% \label{zero-shot_table}
% \end{table}

\begin{table}[h]
\centering
\resizebox{0.85\linewidth}{!}{
\begin{tabular}{l|cccc}
\hline
\textbf{Methods} & \textbf{PSNR$\uparrow$}  & \textbf{SSIM$\uparrow$} & \textbf{SAM$\downarrow$} & \textbf{ERGAS$\downarrow$}  \\
\hline
Bicubic &34.0869 &0.8726 &0.0397 &2.1202\\
SFINet++ & 33.3047 & 0.8679 & 0.0439 & 2.3105\\
Pan-Mamba & 30.5913 & 0.7656 & 0.0524 & 3.1224 \\
\hline 
Ours & \textbf{36.4892} & \textbf{0.9114} & \textbf{0.0299} & \textbf{1.5515}\\
\hline
\end{tabular}}
\caption{Comparison results on the WV2 dataset for zero-shot image super-resolution evaluation.}
\label{zero-shot_table}
\end{table}

\begin{table}[t]
\centering
% \footnotesize
\resizebox{0.65\linewidth}{!}{
\begin{tabular}{c|ccc} 
	\hline
\textbf{Methods} & \textbf{FLOPs (G)} & \textbf{Params (M)}\\\hline
SRPPNN &  21.1059 & 1.7114 \\
INNformer  &  1.3079 & 0.0706\\
FAME &  9.4093 & 0.5766\\
WaveletNet &  7.770 & 1.3230\\
% Mutual \cite{zhou2022mutual} & 12.42 & 697.27 \\
SFINet++ & 1.3112 & 0.0848\\
Pan-Mamba & 3.0088 & 0.1827\\
CFLIHPs & 6.4500 & 0.1314\\
\hline
	Ours& 5.0616 & 0.2453\\
	\hline
\end{tabular}}
\caption{The comparison of computational efficiency.}
\label{computational efficiency}
\end{table}

\subsection{Computational Efficiency}
We have evaluated the FLOPs and the number of parameters of our proposed method, along with other comparative methods, on PAN images with a resolution of $128\times128$ and MS images with a resolution $32\times32$ on a single Nvidia V100 GPU. The results of this evaluation are presented in Table \ref{computational efficiency}. Our proposed method demonstrates 5.0616 G FLOPs and 0.2453 M parameters.

\section{Conclusion}
In conclusion, we present MMMamba, a novel cross-modal in-context fusion framework that pioneers the exploration of the in-context conditioning paradigm in the pan-sharpening domain. Built on the Mamba architecture, MMMamba achieves linear computational complexity and enables efficient bidirectional information flow between PAN and MS modalities. To further strengthen multimodal interactions, we design a multimodal interleaved scanning mechanism that effectively captures complementary characteristics across modalities. Our framework also demonstrates strong generalization capabilities, including zero-shot adaptation to image super-resolution task. Extensive experiments across multiple benchmark datasets consistently validate the superiority of MMMamba over existing SOTA methods. 

\section*{Acknowledgments} 
% \section{Acknowledgments}
This work was supported by the National Natural Science Foundation of China under Grant 82272071, 62271430, 82172073, and 52105126.

\bibliography{aaai2026}

@inproceedings{jenerowicz2016pan,
  title={The pan-sharpening of satellite and UAV imagery for agricultural applications},
  author={Jenerowicz, Agnieszka and Woroszkiewicz, Malgorzata},
  booktitle={Remote Sensing for Agriculture, Ecosystems, and Hydrology XVIII},
  volume={9998},
  pages={565--575},
  year={2016},
  organization={SPIE}
}

@inproceedings{aiazzi2003mtf,
  title={An MTF-based spectral distortion minimizing model for pan-sharpening of very high resolution multispectral images of urban areas},
  author={Aiazzi, B and Alparone, Luciano and Baronti, S and Garzelli, Andrea and Selva, M},
  booktitle={2003 2nd GRSS/ISPRS Joint Workshop on Remote Sensing and Data Fusion over Urban Areas},
  pages={90--94},
  year={2003},
  organization={IEEE}
}

@article{sunuprapto2016evaluation,
  title={Evaluation of pan-sharpening method: applied to artisanal gold mining monitoring in Gunung Pani Forest area},
  author={Sunuprapto, Heri and Danoedoro, Projo and Ritohardoyo, Su},
  journal={Procedia Environmental Sciences},
  volume={33},
  pages={230--238},
  year={2016},
  publisher={Elsevier}
}

@article{mallat2002theory,
  title={A theory for multiresolution signal decomposition: the wavelet representation},
  author={Mallat, Stephane G},
  journal={IEEE transactions on pattern analysis and machine intelligence},
  volume={11},
  number={7},
  pages={674--693},
  year={2002},
  publisher={Ieee}
}

@article{ballester2006variational,
  title={A variational model for P+ XS image fusion},
  author={Ballester, Coloma and Caselles, Vicent and Igual, Laura and Verdera, Joan and Roug{\'e}, Bernard},
  journal={International Journal of Computer Vision},
  volume={69},
  number={1},
  pages={43--58},
  year={2006},
  publisher={Springer}
}

@article{he2023multiscale,
  title={Multiscale dual-domain guidance network for pan-sharpening},
  author={He, Xuanhua and Yan, Keyu and Zhang, Jie and Li, Rui and Xie, Chengjun and Zhou, Man and Hong, Danfeng},
  journal={IEEE Transactions on Geoscience and Remote Sensing},
  volume={61},
  pages={1--13},
  year={2023},
  publisher={IEEE}
}

@inproceedings{li2025freq,
  title={Freq-RWKV: Granularity-Aware Spatial-Frequency Synergy via Dual-Domain Recurrent Scanning for Pan-sharpening},
  author={Li, Xueheng and He, Xuanhua and Hu, Tao and Zhang, Jie and Zhou, Man and Xie, Chengjun and Wang, Yingying and Huang, Bo},
  booktitle={Proceedings of the 33rd ACM International Conference on Multimedia},
  pages={1890--1899},
  year={2025}
}

@inproceedings{meng2025accelerated,
  title={Accelerated Diffusion via High-Low Frequency Decomposition for Pan-Sharpening},
  author={Meng, Ge and Huang, Jingjia and Tu, Jingyan and Wang, Yingying and Lin, Yunlong and Tu, Xiaotong and Huang, Yue and Ding, Xinghao},
  booktitle={Proceedings of the AAAI Conference on Artificial Intelligence},
  volume={39},
  number={6},
  pages={6117--6125},
  year={2025}
}

@article{zhang2024frequency,
  title={Frequency decoupled domain-irrelevant feature learning for pan-sharpening},
  author={Zhang, Jie and Cao, Ke and Yan, Keyu and Lin, Yunlong and He, Xuanhua and Wang, Yingying and Li, Rui and Xie, Chengjun and Zhang, Jun and Zhou, Man},
  journal={IEEE Transactions on Circuits and Systems for Video Technology},
  year={2024},
  publisher={IEEE}
}

@inproceedings{huang2023dp,
  title={Dp-innet: Dual-path implicit neural network for spatial and spectral features fusion in pan-sharpening},
  author={Huang, Jingjia and Meng, Ge and Wang, Yingying and Lin, Yunlong and Huang, Yue and Ding, Xinghao},
  booktitle={Chinese Conference on Pattern Recognition and Computer Vision (PRCV)},
  pages={268--279},
  year={2023},
  organization={Springer}
}

@article{wang2025towards,
  title={Towards Generalizable Pan-sharpening: Conditional Flow-based Learning Guided by Implicit High-frequency Priors},
  author={Wang, Yingying and Zheng, Hui and Li, Feifei and Lin, Yunlong and Fan, Linyu and He, Xuanhua and Huang, Yue and Ding, Xinghao},
  journal={IEEE Transactions on Geoscience and Remote Sensing},
  year={2025},
  publisher={IEEE}
}

@article{wang2025learning,
  title={Learning Diffusion High-Quality Priors for Pan-Sharpening: A Two-Stage Approach With Time-Aware Adapter Fine-Tuning},
  author={Wang, Yingying and Lin, Yunlong and He, Xuanhua and Zheng, Hui and Yan, Keyu and Fan, Linyu and Huang, Yue and Ding, Xinghao},
  journal={IEEE Transactions on Geoscience and Remote Sensing},
  year={2025},
  publisher={IEEE}
}

@article{he2025pan,
  title={Pan-mamba: Effective pan-sharpening with state space model},
  author={He, Xuanhua and Cao, Ke and Zhang, Jie and Yan, Keyu and Wang, Yingying and Li, Rui and Xie, Chengjun and Hong, Danfeng and Zhou, Man},
  journal={Information Fusion},
  volume={115},
  pages={102779},
  year={2025},
  publisher={Elsevier}
}

@inproceedings{esser2024scaling,
  title={Scaling rectified flow transformers for high-resolution image synthesis},
  author={Esser, Patrick and Kulal, Sumith and Blattmann, Andreas and Entezari, Rahim and M{\"u}ller, Jonas and Saini, Harry and Levi, Yam and Lorenz, Dominik and Sauer, Axel and Boesel, Frederic and others},
  booktitle={Forty-first international conference on machine learning},
  year={2024}
}

@article{kwarteng1989extracting,
  title={Extracting spectral contrast in Landsat Thematic Mapper image data using selective principal component analysis},
  author={Kwarteng, P and Chavez, A},
  journal={Photogramm. Eng. Remote Sens},
  volume={55},
  number={1},
  pages={339--348},
  year={1989}
}

@article{gillespie1987color,
  title={Color enhancement of highly correlated images. II. Channel ratio and “chromaticity” transformation techniques},
  author={Gillespie, Alan R and Kahle, Anne B and Walker, Richard E},
  journal={Remote Sensing of Environment},
  volume={22},
  number={3},
  pages={343--365},
  year={1987},
  publisher={Elsevier}
}

@article{schowengerdt1980reconstruction,
  title={Reconstruction of multispatial, multispectral image data using spatial frequency content},
  author={Schowengerdt, Robert A},
  journal={Photogrammetric Engineering and Remote Sensing},
  volume={46},
  number={10},
  pages={1325--1334},
  year={1980}
}

@article{nunez2002multiresolution,
  title={Multiresolution-based image fusion with additive wavelet decomposition},
  author={Nunez, Jorge and Otazu, Xavier and Fors, Octavi and Prades, Albert and Pala, Vicenc and Arbiol, Roman},
  journal={IEEE Transactions on Geoscience and Remote sensing},
  volume={37},
  number={3},
  pages={1204--1211},
  year={2002},
  publisher={IEEE}
}

@article{fasbender2008bayesian,
  title={Bayesian data fusion for adaptable image pansharpening},
  author={Fasbender, Dominique and Radoux, Julien and Bogaert, Patrick},
  journal={IEEE Transactions on Geoscience and Remote Sensing},
  volume={46},
  number={6},
  pages={1847--1857},
  year={2008},
  publisher={IEEE}
}

@article{masi2016pansharpening,
  title={Pansharpening by convolutional neural networks},
  author={Masi, Giuseppe and Cozzolino, Davide and Verdoliva, Luisa and Scarpa, Giuseppe},
  journal={Remote Sensing},
  volume={8},
  number={7},
  pages={594},
  year={2016},
  publisher={MDPI}
}

@inproceedings{yang2017pannet,
  title={PanNet: A deep network architecture for pan-sharpening},
  author={Yang, Junfeng and Fu, Xueyang and Hu, Yuwen and Huang, Yue and Ding, Xinghao and Paisley, John},
  booktitle={Proceedings of the IEEE international conference on computer vision},
  pages={5449--5457},
  year={2017}
}

@article{cai2020super,
  title={Super-resolution-guided progressive pansharpening based on a deep convolutional neural network},
  author={Cai, Jiajun and Huang, Bo},
  journal={IEEE Transactions on Geoscience and Remote Sensing},
  volume={59},
  number={6},
  pages={5206--5220},
  year={2020},
  publisher={IEEE}
}

@inproceedings{wang2023learning,
  title={Learning high-frequency feature enhancement and alignment for pan-sharpening},
  author={Wang, Yingying and Lin, Yunlong and Meng, Ge and Fu, Zhenqi and Dong, Yuhang and Fan, Linyu and Yu, Hedeng and Ding, Xinghao and Huang, Yue},
  booktitle={Proceedings of the 31st ACM International Conference on Multimedia},
  pages={358--367},
  year={2023}
}

@inproceedings{zhou2022panformer,
  title={PanFormer: A transformer based model for pan-sharpening},
  author={Zhou, Huanyu and Liu, Qingjie and Wang, Yunhong},
  booktitle={2022 IEEE international conference on multimedia and expo (ICME)},
  pages={1--6},
  year={2022},
  organization={IEEE}
}

@inproceedings{zhou2022pan,
  title={Pan-sharpening with customized transformer and invertible neural network},
  author={Zhou, Man and Huang, Jie and Fang, Yanchi and Fu, Xueyang and Liu, Aiping},
  booktitle={Proceedings of the AAAI conference on artificial intelligence},
  volume={36},
  number={3},
  pages={3553--3561},
  year={2022}
}

@article{wang2024cross,
  title={Cross-modality interaction network for pan-sharpening},
  author={Wang, Yingying and He, Xuanhua and Dong, Yuhang and Lin, Yunlong and Huang, Yue and Ding, Xinghao},
  journal={IEEE Transactions on Geoscience and Remote Sensing},
  volume={62},
  pages={1--16},
  year={2024},
  publisher={IEEE}
}

@article{gu2021efficiently,
  title={Efficiently modeling long sequences with structured state spaces},
  author={Gu, Albert and Goel, Karan and R{\'e}, Christopher},
  journal={arXiv preprint arXiv:2111.00396},
  year={2021}
}

@article{liu2024vmamba,
  title={Vmamba: Visual state space model},
  author={Liu, Yue and Tian, Yunjie and Zhao, Yuzhong and Yu, Hongtian and Xie, Lingxi and Wang, Yaowei and Ye, Qixiang and Jiao, Jianbin and Liu, Yunfan},
  journal={Advances in neural information processing systems},
  volume={37},
  pages={103031--103063},
  year={2024}
}

@article{zhu2024vision,
  title={Vision mamba: Efficient visual representation learning with bidirectional state space model},
  author={Zhu, Lianghui and Liao, Bencheng and Zhang, Qian and Wang, Xinlong and Liu, Wenyu and Wang, Xinggang},
  journal={arXiv preprint arXiv:2401.09417},
  year={2024}
}

@article{gu2023mamba,
  title={Mamba: Linear-time sequence modeling with selective state spaces},
  author={Gu, Albert and Dao, Tri},
  journal={arXiv preprint arXiv:2312.00752},
  year={2023}
}

@inproceedings{cao2024novel,
  title={A novel state space model with local enhancement and state sharing for image fusion},
  author={Cao, Zihan and Wu, Xiao and Deng, Liang-Jian and Zhong, Yu},
  booktitle={Proceedings of the 32nd ACM International Conference on Multimedia},
  pages={1235--1244},
  year={2024}
}

@article{tan2024ominicontrol,
  title={Ominicontrol: Minimal and universal control for diffusion transformer},
  author={Tan, Zhenxiong and Liu, Songhua and Yang, Xingyi and Xue, Qiaochu and Wang, Xinchao},
  journal={arXiv preprint arXiv:2411.15098},
  year={2024}
}

@article{labs2025flux,
  title={FLUX. 1 Kontext: Flow Matching for In-Context Image Generation and Editing in Latent Space},
  author={Labs, Black Forest and Batifol, Stephen and Blattmann, Andreas and Boesel, Frederic and Consul, Saksham and Diagne, Cyril and Dockhorn, Tim and English, Jack and English, Zion and Esser, Patrick and others},
  journal={arXiv preprint arXiv:2506.15742},
  year={2025}
}

@inproceedings{he2024frequency,
  title={Frequency-adaptive pan-sharpening with mixture of experts},
  author={He, Xuanhua and Yan, Keyu and Li, Rui and Xie, Chengjun and Zhang, Jie and Zhou, Man},
  booktitle={Proceedings of the AAAI Conference on Artificial Intelligence},
  volume={38},
  number={3},
  pages={2121--2129},
  year={2024}
}

@inproceedings{lin2023domain,
  title={Domain-irrelevant feature learning for generalizable pan-sharpening},
  author={Lin, Yunlong and Fu, Zhenqi and Meng, Ge and Wang, Yingying and Dong, Yuhang and Fan, Linyu and Yu, Hedeng and Ding, Xinghao},
  booktitle={Proceedings of the 31st ACM International Conference on Multimedia},
  pages={3287--3296},
  year={2023}
}

@article{hornet,
  title={Hornet: Efficient high-order spatial interactions with recursive gated convolutions},
  author={Rao, Yongming and Zhao, Wenliang and Tang, Yansong and Zhou, Jie and Lim, Ser Nam and Lu, Jiwen},
  journal={Advances in Neural Information Processing Systems},
  volume={35},
  pages={10353--10366},
  year={2022}
}

@article{zhao2016loss,
  title={Loss functions for image restoration with neural networks},
  author={Zhao, Hang and Gallo, Orazio and Frosio, Iuri and Kautz, Jan},
  journal={IEEE Transactions on computational imaging},
  volume={3},
  number={1},
  pages={47--57},
  year={2016},
  publisher={IEEE}
}

@article{liu2000smoothing,
  title={Smoothing filter-based intensity modulation: A spectral preserve image fusion technique for improving spatial details},
  author={Liu, JG},
  journal={International Journal of remote sensing},
  volume={21},
  number={18},
  pages={3461--3472},
  year={2000},
  publisher={Taylor \& Francis}
}

@inproceedings{liao2015two,
  title={Two-stage fusion of thermal hyperspectral and visible RGB image by PCA and guided filter},
  author={Liao, Wenzhi and Huang, Xin and Van Coillie, Frieke and Thoonen, Guy and Pi{\v{z}}urica, Aleksandra and Scheunders, Paul and Philips, Wilfried},
  booktitle={2015 7th Workshop on Hyperspectral Image and Signal Processing: Evolution in Remote Sensing (WHISPERS)},
  pages={1--4},
  year={2015},
  organization={Ieee}
}

@article{wu2023lrtcfpan,
  title={LRTCFPan: Low-rank tensor completion based framework for pansharpening},
  author={Wu, Zhong-Cheng and Huang, Ting-Zhu and Deng, Liang-Jian and Huang, Jie and Chanussot, Jocelyn and Vivone, Gemine},
  journal={IEEE Transactions on Image Processing},
  volume={32},
  pages={1640--1655},
  year={2023},
  publisher={IEEE}
}

@inproceedings{haydn1982application,
  title={Application of the IHS color transform to the processing of multisensor data and image enhancement},
  author={Haydn, R},
  booktitle={Proc. of the International Symposium on Remote Sensing of Arid and Semi-Arid Lands, Cairo, Egypt, 1982},
  year={1982}
}

@article{zhou2024general,
  title={A general spatial-frequency learning framework for multimodal image fusion},
  author={Zhou, Man and Huang, Jie and Yan, Keyu and Hong, Danfeng and Jia, Xiuping and Chanussot, Jocelyn and Li, Chongyi},
  journal={IEEE Transactions on Pattern Analysis and Machine Intelligence},
  year={2024},
  publisher={IEEE}
}

@article{WaveletNet,
  title={Pan-sharpening with wavelet-enhanced high-frequency information},
  author={Zhang, Jie and He, Xuanhua and Yan, Keyu and Cao, Ke and Li, Rui and Xie, Chengjun and Zhou, Man and Hong, Danfeng},
  journal={IEEE Transactions on Geoscience and Remote Sensing},
  volume={62},
  pages={1--14},
  year={2024},
  publisher={IEEE}
}

@inproceedings{meng2024progressive,
  title={Progressive high-frequency reconstruction for pan-sharpening with implicit neural representation},
  author={Meng, Ge and Huang, Jingjia and Wang, Yingying and Fu, Zhenqi and Ding, Xinghao and Huang, Yue},
  booktitle={Proceedings of the AAAI Conference on Artificial Intelligence},
  volume={38},
  number={5},
  pages={4189--4197},
  year={2024}
}

@inproceedings{hou2024linearly,
  title={Linearly-evolved transformer for pan-sharpening},
  author={Hou, Junming and Cao, Zihan and Zheng, Naishan and Li, Xuan and Chen, Xiaoyu and Liu, Xinyang and Cong, Xiaofeng and Hong, Danfeng and Zhou, Man},
  booktitle={Proceedings of the 32nd ACM international conference on multimedia},
  pages={1486--1494},
  year={2024}
}

@article{wu2025fully,
  title={Fully-connected transformer for multi-source image fusion},
  author={Wu, Xiao and Cao, Zi-Han and Huang, Ting-Zhu and Deng, Liang-Jian and Chanussot, Jocelyn and Vivone, Gemine},
  journal={IEEE Transactions on Pattern Analysis and Machine Intelligence},
  volume={47},
  number={3},
  pages={2071--2088},
  year={2025},
  publisher={IEEE}
}

@inproceedings{hou2023bidomain,
  title={Bidomain modeling paradigm for pansharpening},
  author={Hou, Junming and Cao, Qi and Ran, Ran and Liu, Che and Li, Junling and Deng, Liang-jian},
  booktitle={Proceedings of the 31st ACM international conference on multimedia},
  pages={347--357},
  year={2023}
}
%newpage
\end{document}